\documentclass[10pt,twocolumn,letterpaper]{article}

\usepackage[pagenumbers]{cvpr} 

\usepackage{graphicx}
\usepackage{amsmath}
\usepackage{amssymb}
\usepackage{booktabs}
\usepackage{enumitem}
\usepackage{multirow}

\usepackage{mathtools}

\usepackage[accsupp]{axessibility}

\usepackage{color}

\usepackage{bibentry}
\definecolor{Blue}{RGB}{0,0,255}
\definecolor{Red}{RGB}{255,0,0}
\usepackage[pagebackref,breaklinks,colorlinks]{hyperref}

\usepackage[capitalize]{cleveref}
\crefname{section}{Sec.}{Secs.}
\Crefname{section}{Section}{Sections}
\Crefname{table}{Table}{Tables}
\crefname{table}{Tab.}{Tabs.}

\def\cvprPaperID{6924} 
\def\confName{CVPR}
\def\confYear{2022}

\begin{document}

\title{SC$^2$-PCR: A Second Order Spatial Compatibility for Efficient and Robust \\ Point Cloud Registration}

\author{Zhi Chen\textsuperscript{\rm 1} \and Kun Sun\textsuperscript{\rm 2} \and Fan Yang\textsuperscript{\rm 1} \and Wenbing Tao\textsuperscript{\rm 1}\thanks{Corresponding author.} \and 
	\textsuperscript{\rm 1}National Key Laboratory of Science and Technology on Multi-spectral Information Processing,\\
	School of Artifical Intelligence and Automation, Huazhong University of Science and Technology, China \\
	\textsuperscript{\rm 2}Hubei Key Laboratory of Intelligent Geo-Information Processing,\\ School of Computer Science, 
	China University of Geosciences, China \\	
	{\tt\small \{z\_chen, fanyang, wenbingtao\}@hust.edu.cn; sunkun@cug.edu.cn} \\
}
\maketitle

\begin{abstract}

In this paper, we present a second order spatial compatibility (SC$^2$) measure based method for efficient and robust point cloud registration (PCR), called SC$^2$-PCR \footnote{Code will be available at \url{https://github.com/ZhiChen902/SC2-PCR}.}. Firstly, we propose a second order spatial compatibility (SC$^2$) measure to compute the similarity between correspondences. It considers the global compatibility instead of local consistency, allowing for more distinctive clustering between inliers and outliers at early stage. Based on this measure, our registration pipeline employs a global spectral technique to find some reliable seeds from the initial correspondences. Then we design a two-stage strategy to expand each seed to a consensus set based on the SC$^2$ measure matrix. Finally, we feed each consensus set to a weighted SVD algorithm to generate a candidate rigid transformation and select the best model as the final result. Our method can guarantee to find a certain number of outlier-free consensus sets using fewer samplings, making the model estimation more efficient and robust. In addition, the proposed SC$^2$ measure is general and can be easily plugged into deep learning based frameworks. Extensive experiments are carried out to investigate the performance of our method. 
\end{abstract}

\section{Introduction}
\label{sec:intro}
The alignment of two 3D scans of the same scene, known as Point Cloud Registration(PCR), plays an important role in areas such as Simultaneous Localization and Mapping (SLAM) \cite{bailey2006simultaneous,durrant2006simultaneous,montemerlo2002fastslam}, augmented reality  \cite{azuma1997survey,billinghurst2015survey} and robotics applications \cite{kostavelis2015semantic}. A canonical solution first establishes feature correspondences and then estimates the 3D rotation and translation that can best align the shared parts. However, due to challenges such as partial overlap or feature ambiguity, model estimation is prone to outliers in the correspondences, leading to inaccurate or wrong alignment.

\begin{figure}[t]
	\centering
	\includegraphics[width=1\columnwidth]{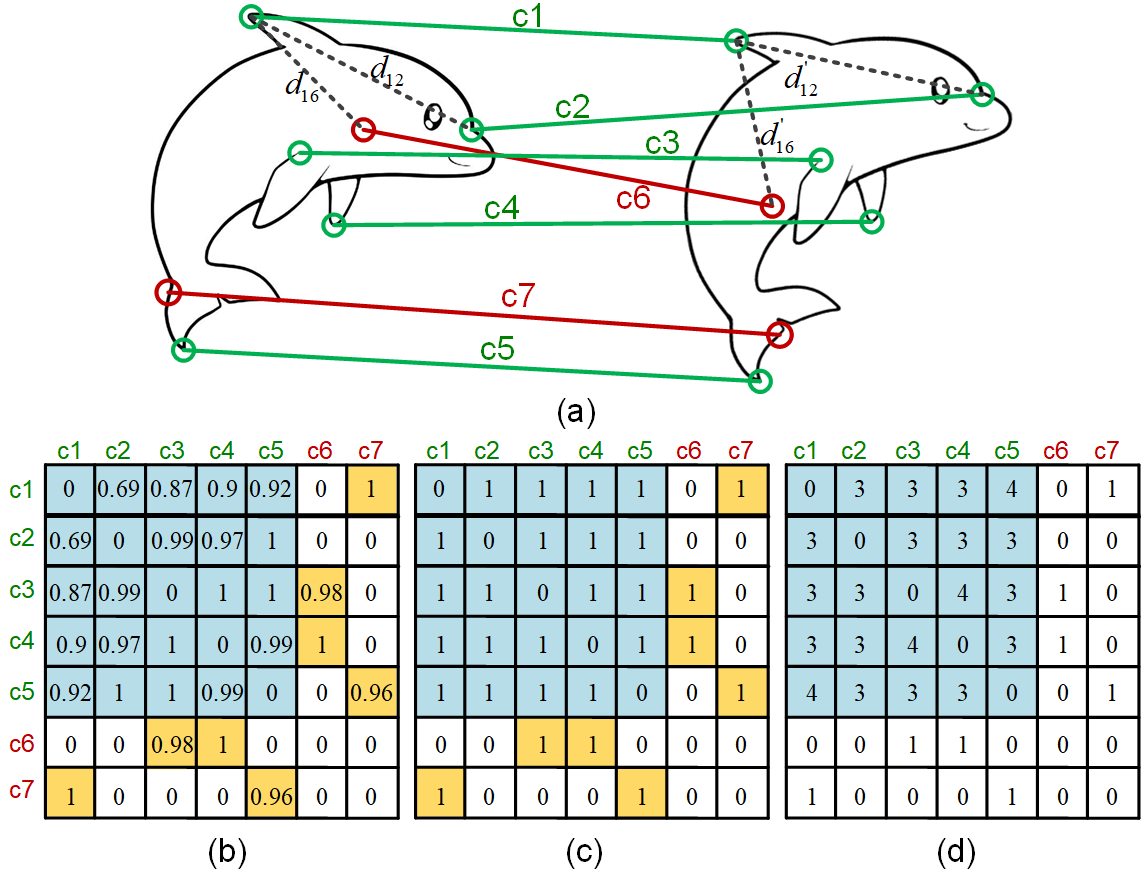}
	\caption{\textbf{(a):} A toy example in which red and green line segments represent outliers and inliers, respectively. \textbf{(b):} The first order compatibility matrix of (a). As highlighted by yellow, the outliers have very high compatibility scores with some inliers. \textbf{(c):} A binarized compatibility matrix of (b) after thresholding. \textbf{(d):} The proposed second order compatibility matrix of (a). By contrast, the values in the rows and columns of the outliers are small.} \label{fig:heatmap}
	\vspace{-5mm}
\end{figure}

RANSAC \cite{fischler1981random} pioneers the iterative sampling strategy for model estimation. However, it needs more time to converge or sometimes there is no guarantee of an accurate solution due to massive outliers. Spatial Compatibility (SC) \cite{bai2021pointdsc,quan2020compatibility,lee2021deep,yang2021sac} is a widely used similarity measure for boosting the robustness and efficiency of the rigid transformation estimation. It assumes that two correspondences will have higher score if the difference of spatial distance between them, e.g. $|d_{12}-d_{12}^{'}|$ or $|d_{16}-d_{16}^{'}|$ in Fig. \ref{fig:heatmap}(a), is minor. Thus, sampling from compatible correspondences increases the probability of getting inliers. However, such kind of first order metric still suffers from outliers due to locality and ambiguity. Fig. \ref{fig:heatmap}(b) is the spatial compatibility matrix of the correspondences in Fig. \ref{fig:heatmap}(a). As we can see from the yellow cells, $c_6$ and $c_7$ are outliers but they show high compatibility scores with some inliers by chance. As a result, the outliers would be inevitably involved in the model estimation process, leading to performance deterioration.

In this paper, we propose a new global measure of the similarity between two correspondences. Specifically, we first binarize the spatial compatibility matrix into the hard form, as shown in Fig. \ref{fig:heatmap} (c). Then,
for two correspondences which are compatible, we compute the number of their commonly compatible correspondences in the global set. That is, we compute the number of correspondences that are simultaneously compatible with both of them as the new similarity between them. The globally common compatibility is set to 0 for any two incompatible correspondences. Therefore, the similarity of two inliers is at least the number of inliers excluding themselves among all the correspondences. However, the outliers do not have such good properties. A toy example is shown in Fig. \ref{fig:heatmap}. There are five inliers \{c1,c2, c3, c4, c5\} and two outliers \{c6, c7\} in Fig. \ref{fig:heatmap} (a). From Fig. \ref{fig:heatmap} (b) and Fig. \ref{fig:heatmap} (c), we can see that the outliers  $c_6$ and $c_7$ are compatible with some inliers, and the inliers are compatible with each other. But from Fig. \ref{fig:heatmap} (d), we can see that the similarities between any two inliers are large while the similarities between the outliers with the other correspondences are small. To be specific, in Fig. \ref{fig:heatmap} (d), the similarities within the inliers \{c1, c2, c3, c4, c5\} are no less than 3, while the similarities related with the outliers \{c6, c7\} are no more than 1. Therefore, the global compatibility matrix in Fig. \ref{fig:heatmap} (d)  can better distinguish inliers from outliers.  
Since the new measure can be expressed as the matrix product of the traditional first order metric (See Eq. \ref{global_compatibility}), we name it as the second order spatial compatibility (SC$^2$).

The proposed second order spatial compatibility measure SC$^2$ has several advantages. 1) The inliers are much easier distinguished from the outliers. Suppose we have $m$ inliers in $n$ correspondences. The scores between any two inliers would be no less than $m$-2. However, it’s difficult for an outlier to be simultaneously compatible with multiple correspondences and the score for it will be much smaller. 2) The traditional algorithms such as RANSAC and its variants \cite{fischler1981random,torr2002napsac,ni2009groupsac,chum2005matching} need a large number of randomly samplings to find an outlier-free set for robust model estimation. However, based on the proposed  SC$^2$ matrix, for each row vector corresponding to an inlier,  we can easily find an outlier-free set by selecting the top $k$ correspondences with the highest scores. In this way, the $m$ valid samplings can be obtained by traversing all the $n$ rows of the SC$^2$ matrix. Therefore, we can ensure $m$ reliable model estimations by only  $n$ samplings, which makes the model estimation more efficient and robust. 
3) We theoretically prove that the SC$^2$ matrix significantly reduces the probability of wrong sampling from a probabilistic view.
We define an error event, in which the score between two inliers is smaller than that between an inlier and an outlier. By computing the probability distributions of this event for both the traditional first order metric and our second order metric, the SC$^2$ matrix is much more robust to obtain reliable sampling (see Fig. \ref{fig:probability}).

Based on the SC$^2$ measure, we design a full pipeline for point cloud registration, called SC$^2$-PCR. Following \cite{bai2021pointdsc,cavalli2020handcrafted,sun2019guide}, it first selects several seeds that are likely to be inliers. Then we select a consensus set for each seed by finding those having the highest SC$^2$ scores with it. In order to further exclude outliers, a two-stage sampling strategy is carried out in a coarse-to-fine manner. Finally, we use the weighted SVD to estimate a tentative model for each seed and select the best one as the final output. In a nutshell,
this paper distinguishes itself from existing methods in the following aspects. 
\begin{itemize}[itemsep=2pt,topsep=0pt,parsep=0pt]
	\item A second order spatial compatibility metric called SC$^2$ is proposed. We theoretically prove that SC$^2$ significantly reduces the probability of an outlier being involved in the consensus set. Since the proposed method encodes richer information beyond the first order metric, it enhances the robustness against outliers. 
	\item Compared with state-of-the-art deep learning methods such as \cite{pais20203dregnet,choy2020deep,bai2021pointdsc,lee2021deep}, our method is a light weighted solution that does not need training. It shows no bias across different datasets and generalizes well on various scenarios, which is also shown in the experiments.
	\item The proposed method is general. Although we implement it in a handcrafted fashion, it could be easily plugged into other deep learning frameworks such as PointDSC \cite{bai2021pointdsc}. We show in the experiment that PointDSC produces better results when combined with our method.
\end{itemize}

\section{Related Work}
\textbf{3D Feature Matching.}
The widely used Iterative Closest Point \cite{besl1992method} and its variants \cite{rusinkiewicz2001efficient,segal2009generalized,bouaziz2013sparse} establish correspondences by searching the nearest neighbor in coordinate space. Instead of using the distance in coordinate space, local feature descriptors aim to build 3D matches in feature space. Hand-crafted descriptors represent local feature by encoding spatial distribution histogram \cite{johnson1999using,frome2004recognizing,tombari2010unique,guo2015novel}, geometric histogram \cite{chen20073d,rusu2008aligning,rusu2009fast} or other attributes \cite{zaharescu2009surface}. Recently, the deep learning techniques are also introduced to learn 3D local descriptors. The pioneering 3DMatch \cite{zeng20173dmatch} builds a Siamese Network for extracting local descriptors. A number of recent networks try to boost the performance by designing rotation invariant modules \cite{deng2018ppfnet, yang2018foldingnet, deng2018ppf, ao2021spinnet}, fully convolution modules \cite{choy2019fully}, feature detection modules \cite{bai2020d3feat,yew20183dfeat}, coarse-to-fine modules \cite{yu2021cofinet,qin2022geometric} and overlapping learning modules \cite{huang2021predator,xu2021omnet}. Although these methods achieve remarkable performance improvement, it can hardly establish a totally outlier-free correspondence set.

\textbf{Traditional Model Fitting.} The model fitting methods estimate the geometric model from a noise correspondence set. The classic RANSAC \cite{fischler1981random} provides the most commonly adopted generation-and-verification pipeline for robust removing outliers. Many of its variants \cite{torr2002napsac,ni2009groupsac,chum2005matching} introduce new sampling strategies and local optimization \cite{chum2005matching} to accelerate the estimation or boost the robustness. More recently, Graph-cut RANSAC \cite{barath2018graph} introduces the graphcut technique to better performing local optimization. Magsac \cite{barath2019magsac} proposes a $\sigma$-consensus to build a threshold-free method for RANSAC. For specific 3D model-fitting, FGR \cite{zhou2016fast} uses the Geman-McClure cost function and estimates the model through graduated non-convexity optimization. TEASER \cite{yang2020teaser} reformulates the registration problem using a Truncated Least Squares (TLS) cost and solving it by a general graph-theoretic framework.

\textbf{Learning based Model Fitting.} Recent works also adopt deep learning techniques, which were first studied in 2D matching area, to model fitting tasks. The 2D correspondence selection network CN-Net \cite{moo2018learning} and its variants \cite{brachmann2019neural,zhao2019nm,liu2021learnable,sun2020acne,zhang2019learning,chen2021learning,zhao2021progressive,chen2021cascade} formulate the model fitting as combination of a correspondence classification module and a model estimation module. Recent attempts \cite{pais20203dregnet,choy2020deep,bai2021pointdsc,lee2021deep,chen2021detarnet} also introduce deep learning network for 3D correspondence pruning. 3DRegNet \cite{pais20203dregnet} reformulates the CN-Net \cite{moo2018learning} into 3D form and designs a regression module to solve rigid transformaton. DGR \cite{choy2020deep} introduces the fully convolution to better capture global context. PointDSC \cite{bai2021pointdsc} develops a spatial consistency based non-local module and a Neural Spectral matching
to accelerate the model generation and selection. DetarNet \cite{chen2021detarnet} presents decoupling solutions for translation and rotation.
DHVR \cite{lee2021deep} exploits the deep hough voting to identify the consensus from the Hough space, so as to predict the final transformation.

\begin{figure*}[t]
	\centering
	\includegraphics[width=2\columnwidth]{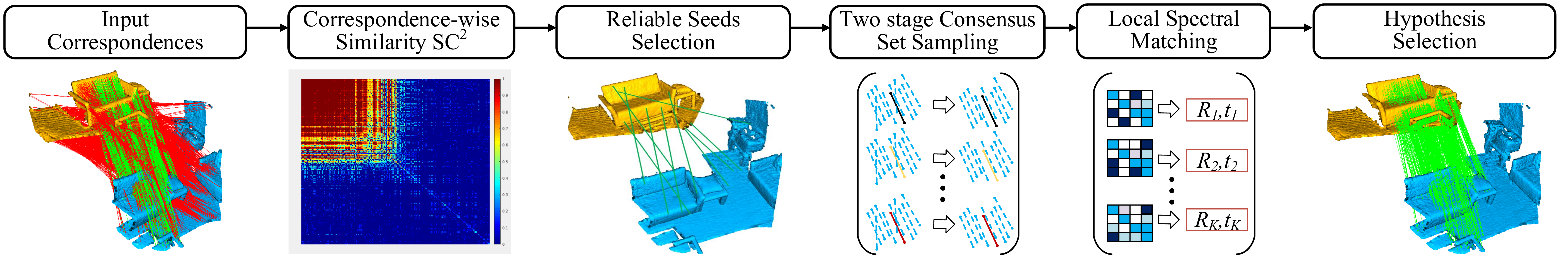}
	\caption{Pipeline of our method. \textbf{1.} Computing correspondence-wise second order spatial compatibility measure. \textbf{2.} Selecting some reliable correspondences as seeds. \textbf{3.} Performing the two-stage sampling around each seed. \textbf{4.} Performing local spectral matching to generate an estimation of $R$ and $t$ for each seed. \textbf{5.} Selecting the best estimation as final result.}
	\vspace{-4mm}
	\label{pipeline}
\end{figure*}

\section{Method}
\subsection{Problem Formulation}
Given two point clouds to be aligned: source point cloud $X=\{x_{i} \in \mathbb{R}^{3} \; | \; i = 1, ..., N_{x}\}$ and target point cloud $Y = \{y_{j} \in \mathbb{R}^{3} \; | \; j = 1, ..., N_{y}\}$, we first extract local features for both of them. Then, for each point in the source point cloud, we find its nearest neighbor in the feature space among the target points to form $N$ pairs of putative correspondences. The proposed method estimates the rigid transformation between the two point clouds, i.e., the rotation matrices ($R \in \mathbb{R}^{3 \times 3})$ and translation vectors ($t \in \mathbb{R}^{3}$). The pipeline of proposed method is shown in Fig. \ref{pipeline}.

\subsection{Second Order Spatial Compatibility}\label{step0}

To analyze the effectiveness of a metric for sampling, we first define the probability of an ambiguity event as:
\begin{equation}
	\begin{aligned}
		{\rm P}_{am}(M) = {\rm P}(M_{in,out} > M_{in,in}),
	\end{aligned}\label{ambiguity_P}
\end{equation}
where $M$ is a specific metric for measuring correspondence-wise similarity. $\rm P(Z)$ is the probability of an event $\rm Z$ (For convenience we use this notation in the following part). $M_{in,out}$ is the similarity between an inlier and an outlier, while $M_{in,in}$ is that between two inliers.
When $M_{in,out} > M_{in,in}$, outlier is the near neighbor of inlier, so the metric based sampling tends to fail. 
So the lower this probability is, the more robust the metric based sampling will be.

\begin{figure}[t]
	\centering
	\includegraphics[width=1\columnwidth]{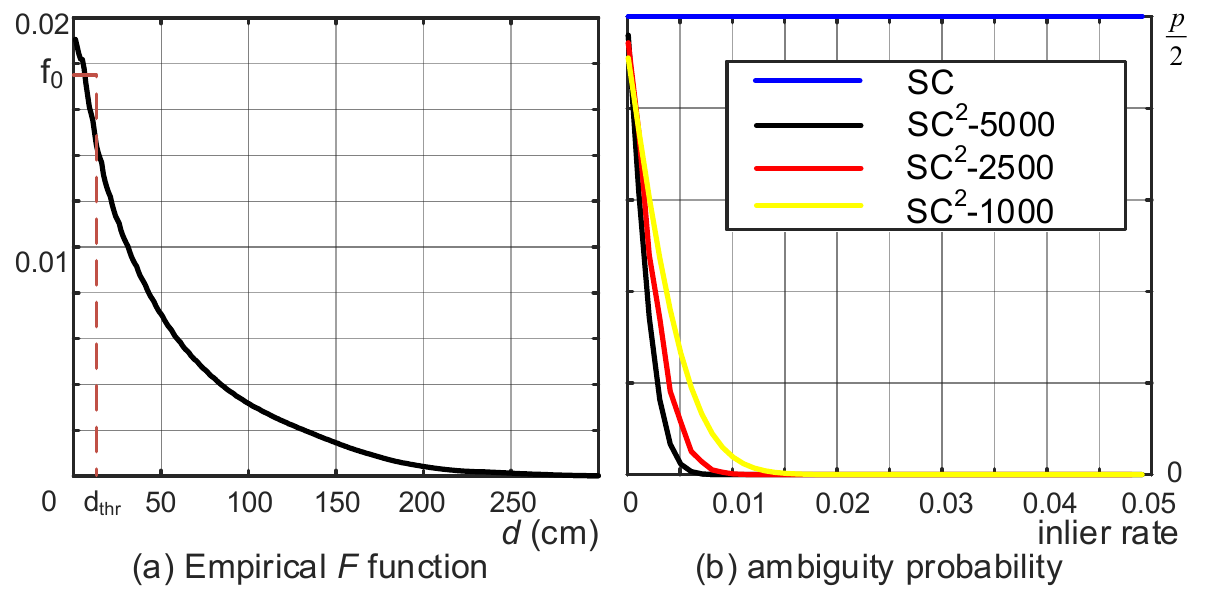}
	\caption{(a) The empirical probability density function ($F$) of $d_{in,out}$ and $d_{out,out}$. 
	(b) The ambiguity probability. SC is spatial compatibility measure. SC$^{2}$-$N$ ($N$ = 5000, 2500, 1000) is the second order spatial compatibility measure with $N$ correspondences.}\label{fig:probability}
	\vspace{-4mm}
\end{figure}

We first introduce the commonly used first order spatial compatibility ($SC$) measure \cite{leordeanu2005spectral,quan2020compatibility,bai2021pointdsc,lee2021deep}. The $SC$ measure between correspondence $i$ and $j$ is defined as follows:
\begin{equation}
	\begin{aligned}
		SC_{ij} = \phi(d_{ij}), d_{ij}=|d(x_i,x_j)-d(y_i,y_j)|
	\end{aligned}\label{spatial_consistency}
\end{equation}
in which ($x_{i}, y_{i}$) and ($x_{j}, y_{j}$) are the matched points of correspondences $i$ and $j$. $\phi(\cdot)$ is a 
monotonically decreasing kernel function. $d(\cdot, \cdot)$ is the Euclidean distance.
As shown in Fig. \ref{fig:heatmap}, the distance difference between two inliers $d_{in,in}$ should be equal to 0 due to the length consistency of rigid transformation.
However, because of the noise introduced by data acquisition and point cloud downsampling, $d_{in,in}$ is not exactly equal to 0,  but less than a threshold $d_{thr}$. For convenience, we assume $d_{in,in}$ is uniformly distributed over $d_{thr}$ and get the probability density function (PDF) of the distance difference between two inliers as follows:
\begin{equation}
	\begin{aligned}
		{\rm PDF}_{in,in}(l) = {1}/{d_{thr}}, 0 \leq l \leq d_{thr}.
	\end{aligned}\label{PDF_in_in}
\end{equation}
Differently, there is no related constraint between two outliers or an inlier and an outlier due to the random distribution of outliers. We consider the distance difference between two unrelated points to be identically distributed and assume the probability density function (PDF) as $F(\cdot)$:
\begin{equation}
	\begin{aligned}
		{\rm PDF}_{in,out}(l) = F(l), {\rm PDF}_{out,out}(l) = F(l); 0 \leq l \leq d_r
	\end{aligned}\label{PDF_in_out}
\end{equation}
where $d_{r}$ is the range of $d_{in,out}$ and $d_{out,out}$. An empirical $F$ function on 3DMatch dataset is presented in Fig \ref{fig:probability} (a). Obviously, $d_{r}$ is much greater than $d_{thr}$. So we can approximate that $F(l)$ is a constant within $(0, d_{thr})$ as follows:
\begin{equation}
	\begin{aligned}
		F(l) = f_{0}, 0 \leq l \leq d_{thr}
	\end{aligned}\label{F0}.
\end{equation}
Next, we consider the ambiguity probability of $SC$ as Eq. (\ref{ambiguity_P}), i.e. , ${\rm P}(SC_{in,out} > SC_{in,in})$. 
According to Eq. (\ref{spatial_consistency}), (\ref{PDF_in_in}), (\ref{PDF_in_out}) and (\ref{F0}), it can be computed as follows:
\begin{equation}
	\begin{aligned}
		&{\rm P}(SC_{in,out} > SC_{in,in}) = {\rm P}(d_{in,out} < d_{in,in}) \\
		=& \int_{0}^{d_{thr}} \int_{0}^{l} {\rm PDF}_{in,in}(l) \cdot {\rm PDF}_{in,out}(x)dxdl \\
		=& \int_{0}^{d_{thr}} \int_{0}^{l} \frac{1}{d_{thr}}  \cdot f_0 dxdl =  \frac{d_{thr} \cdot f_0}{2}
	\end{aligned}\label{ambiguity_SC}.
\end{equation}
Taking the 3DMatch \cite{zeng20173dmatch} dataset as an example. Following \cite{bai2021pointdsc}, we set $d_{thr}$ = 10cm, then the ambiguity probability of $SC$ measure is about 0.1 as shown in Fig. \ref{fig:probability} (a). Considering the amount of outliers might be large, the number of mistakes is not negligible even at this probability.

Next, we describe the proposed second order spatial compatibility measure ($SC^{2} \in \mathbb{R}^{N \times N}$). 
Specifically, we first build a hard compatibility matrix $C$ ($C \in \mathbb{R}^{N \times N}$):
\begin{equation}
	\begin{aligned}
		C_{ij} = \left\{\begin{matrix}
1; d_{ij} \leq d_{thr},\\ 
0; d_{ij} > d_{thr}.
\end{matrix}\right.
	\end{aligned}\label{compatibility_matrix}
\end{equation}
$C$ considers that two correspondences satisfying length consistency are compatible ($C_{i,j}$ is set to 0 when $i=j$). Then, $SC^{2}_{ij}$ counts the number of common compatibility correspondences of $i$ and $j$ when they are compatible, as follows:
\begin{equation}
	\begin{aligned}
		SC^{2}_{ij} = C_{ij} \cdot \sum_{k=1}^{N}C_{ik}  \cdot C_{kj}.
	\end{aligned}\label{global_compatibility}
\end{equation}
Similarly, we analyze the ambiguity probability of this measure, i.e., ${\rm P}(SC^{2}_{in,out} > SC^{2}_{in,in})$. Suppose there are $N$ pairs of correspondences and the inlier ratio is $\alpha$. Then, we can prove that ${\rm P}(SC^{2}_{in,out} > SC^{2}_{in,in})$ can be expressed as (\textit{The derivation is presented in the supplement materials}):
\begin{equation}
	\begin{aligned}
		{\rm P}(SC^{2}_{in,out} > SC^{2}_{in,in}) &=  p \cdot {\rm P}(X > (N \cdot \alpha - 2)), \\
		X \sim S((N \alpha - 1) p + (N (&1 - \alpha) - 1) p^{2}, N (1 - \alpha) p^{2}), \\
		p &= d_{thr} \cdot f_0,
\end{aligned}\label{Ambiguity_SC2}
\end{equation}
where $S(\cdot, \cdot)$ is the Skellam distribution \cite{irwin1937frequency,karlis2003analysis,karlis2006bayesian}. According to the properties of Skellam distribution, the value of ${\rm P}(SC^{2}_{in,out} > SC^{2}_{in,in})$  is going to approach 0 very quickly as $\alpha$ increases. 
In order to make a clearer comparison between the proposed SC$^{2}$ measure and SC measure, we plot the curves of ambiguity probability for both of them according to Eq. (\ref{ambiguity_SC}) and (\ref{Ambiguity_SC2}). 
As shown in Fig. \ref{fig:probability} (b), the ambiguity probability of the proposed SC$^{2}$ measure is significantly lower than the SC measure, even when the inlier rate is close to 0. It shows that using SC$^{2}$ measure as guidance for sampling is easier to obtain an outlier-free set. When the inliers rate reaches 1\%, the ambiguity probability of SC$^2$ measure is close to 0, which ensures a robust sampling anti the data with low inlier rate. 

\subsection{Reliable Seed Selection}\label{step1}
As mentioned above, there are high similarities  
between inlier correspondences by the proposed 
SC$^{2}$ measure. Then, as long as we find an inlier correspondence, we can construct a consensus set by finding its $k$ nearest neighbors in the metric space. Obviously, traversing all the correspondences must find an inlier, but it is not necessary. We only need to pick some reliable points called seed points to accelerate the registration process. 
We perform spectral matching technique \cite{leordeanu2005spectral} to select seed points. Specifically, we first build the similarity matrix for all of the correspondences and normalize the value in the matrix to 0-1, following \cite{leordeanu2005spectral}. Then, following \cite{leordeanu2005spectral,bai2021pointdsc}, the association of each correspondence with the leading eigen-vector is adopted as the confidence for this correspondence. The leading eigen-vetctor is solved by power iteration algorithm \cite{mises1929praktische}. In order to ensure an even distribution of seed points, the correspondences with local maximum confidence score within its neighborhood with radius $R$ are selected. The number of seed points ($N_s$) is determined by a proportion of the number of whole correspondences.

\subsection{Two-stage Consensus Set Sampling}\label{step2}
As some seed points are selected, we extend each of them into a consensus set. 
We adopt a two stage selection strategy to perform a coarse-to-fine sampling.
In the first stage, we select $K_{1}$ correspondences for each seed by finding its top-$K_{1}$ neighbors in the SC$^2$ measure space. As mentioned before, the ambiguity probability ${\rm P}(SC^{2}_{in,out} > SC^{2}_{in,in})$ is very small. Thus, when a seed is an inlier correspondences, the consensus set also mainly contains inliers. Meanwhile, the similarity expressed by SC$^2$ measure focuses on global information instead of local consistency.
Therefore, the neighbors selected in the SC$^2$ measure space are distributed more evenly rather than clustered together, which benefits the estimation of rigid transformation \cite{bai2021pointdsc}.

The second stage sampling operation is adopted to further filter potential outliers in the set obtained in the first stage.
The SC$^2$ matrices are reconstructed within each set produced by the first stage instead of the whole set. We select top-$K_{2}$ ($K_{2} < K_{1}$) correspondences of the seed by the newly constructed local SC$^2$ matrices. As shown in Fig. \ref{fig:probability} (b), since the higher inlier rate ensures a lower ambiguity probability, so that the potential outliers can also be further pruned.
Note that we only discussed the case that the seed point is inlier. In fact, when the seed point is an outlier, it can also form a local consistency, especially when there are aggregated false matching in the correspondence set. We encourage these sets to also generate hypothesis and filter them at the final hypothesis selection step (Section $\ref{step4}$) rather than at the early stage. In this way, we can avoid some correct assumptions being filtered out early.

\subsection{Local Spectral Matching}\label{step3}
In this step, we perform the weighted SVD \cite{arun1987least} on the consensus set to generate an estimation of rigid transformation for each seed. Although the previous proposed sampling strategy can obtain outlier-free correspondence set, we find that the weighted SVD achieves better performance than treating all corrrespondences equally. This may be because that the inliers still have different degree of noises. So correspondences with bigger noise should have smaller weights when estimating rigid transformation. Traditional spectral matching \cite{leordeanu2005spectral} method analyzes the SC matrix to assign weight for each correspondence, which is effected by ambiguity problem \cite{bai2021pointdsc}. 
Since the proposed SC$^2$ measure is more robust against ambiguity, we also replace the SC matrix with SC$^2$ measure. 
In order to facilitate matrix analysis, we convert the SC$^2$ measure into soft form ($\tilde{SC^2}$) as follows:
\begin{equation}
	\begin{aligned}
		\tilde{SC^2} &= \tilde{C} \cdot (\tilde{C} \times \tilde{C}), \\
		\tilde{C}_{ij} = {\rm ReLU}(1 - & d_{ij}^{2} / d_{thr}^2), (1 \leq i \leq K_{2}, 1 \leq j \leq K_{2})
	\end{aligned}\label{soft_global_comp}
\end{equation}
where $\cdot$ is Hadamard product and $\times$ is matrix product.
Then we conduct local spectral decomposition on $\tilde{SC^2}$ to obtain a weight $w_{i}$ for correspondences $i$. 
Finally, the rotation $R_{k}$ and translation $t_{k}$ of seed $k$ are computed by performing weighted SVD \cite{choy2020deep} within its consensus set.

\subsection{Hypothesis Selection}\label{step4}
In final step, we select the best estimation over the rigid transformations produced by all the consensus sets. 
We use the same criteria as RANSAC \cite{fischler1981random}, i.e. inlier count, to select the final estimation. Specifically, for the estimation of $k$-th seed $R_{k}$ and $t_{k}$, we count the number of correspondences that satisfy the constraints of $R_{k}$ and $t_{k}$ by a predefined error threshold ($\tau$) as follows:
\begin{equation}
	\begin{aligned}
		{\rm count}_{k} = \sum_{i = 1}^{N} [||R_{k} x_{i} + t_{k}|| < \tau],
	\end{aligned}\label{count}
\end{equation}
where $[\cdot]$ is Iverson bracket.
The $R_{k}$ and $t_{k}$ with the highest inlier count are selected as the final results.

\section{Experiment}
\begin{table*}[t]
	\centering%
	\scriptsize
	\renewcommand\tabcolsep{4.0pt}
	\begin{tabular}{cc|cccccc|cccccc|c}
		
		\hline
		& & \multicolumn{6}{c|}{FPFH (traditional descriptor)} & \multicolumn{6}{c|}{FCGF (learning based descriptor)} \\ 
		& & RR(\%$\uparrow$) & RE($\circ$$\downarrow$) & TE(cm$\downarrow$) & IP($\%$$\uparrow$) &  IR($\%$$\uparrow$) & F1($\%$$\uparrow$) & RR($\%$$\uparrow$) & RE($\circ$$\downarrow$) & TE(cm$\downarrow$) & IP($\%$$\uparrow$) &  IR($\%$$\uparrow$) & F1($\%$$\uparrow$) & Time(s) \\ \hline
		\multirow{7}{0.04cm}{\rotatebox{90}{learning}} & DCP* \cite{wang2019deep} & 3.22 & 8.42 & 21.40  & - & - & - & 3.22 & 8.42 & 21.40  & - & - & - & 0.07\\
		& PointNetLK* \cite{aoki2019pointnetlk} & 1.61 & 8.04  & 21.30 & - & - & - & 1.61 & 8.04 & 21.30 & - & - & - & 0.12 \\
		& 3DRegNet \cite{pais20203dregnet} & 26.31 & 3.75 & 9.60 & 28.21 & 8.90 & 11.63 & 77.76 & 2.74 & 8.13 & 67.34 & 56.28 & 58.33 & \textcolor{Blue}{0.05}\\
		& DGR \cite{choy2020deep} & 32.84 & 2.45 & 7.53 & 29.51 & 16.78 & 21.35 & 88.85 & 2.28 & 7.02 & 68.51 & 79.92 & 73.15 & 1.53\\
		
		& DHVR \cite{lee2021deep} & 67.10 & 2.78 & 7.84 & 60.19 & 64.90 & 62.11 & 91.93 & \textcolor{Blue}{2.25} & 7.08 & \textcolor{Blue}{80.20} & 78.15 & 78.98 & 3.92\\
		& DHVR-Origin \cite{lee2021deep} & \textcolor{Blue}{80.22} & \textcolor{Blue}{2.06} & 6.87 & & & & 91.40 & \textcolor{Red}{2.08} & 6.61 & & & & 0.46\\
		& PointDSC \cite{bai2021pointdsc} & 77.57 & \textcolor{Red}{2.03} & \textcolor{Red}{6.38} & 68.45 & \textcolor{Blue}{71.56} & \textcolor{Blue}{69.75} & \textcolor{Blue}{92.85} & \textcolor{Red}{2.08} & \textcolor{Red}{6.51} & 78.91 & 86.23 & \textcolor{Blue}{82.12} & 0.10\\ \hline

		\multirow{10}{0.04cm}{\rotatebox{90}{traditional}} & SM \cite{leordeanu2005spectral} & 55.88 & 2.94 & 8.15 & 47.96 & 70.69 & 50.70 & 86.57 & 2.29 & 7.07 & 81.44 & 38.36 & 48.21 & \textcolor{Red}{0.03}\\
		& ICP* \cite{besl1992method} & 5.79 & 7.93 & 17.59 & - & - & - & 5.79 & 7.93 & 17.59 & - & - & - & 0.25\\
		& FGR \cite{zhou2016fast} & 40.91 & 4.96 & 10.25 & 6.84 & 38.90 & 11.23 & 78.93 & 2.90 & 8.41 & 25.63 & 53.90 & 33.58 & 0.89 \\
		& TEASER \cite{yang2020teaser} & 75.48 & 2.48 & 7.31 & \textcolor{Red}{73.01} & 62.63 & 66.93 & 85.77 & 2.73 & 8.66 & \textcolor{Red}{82.43} & 68.08 & 73.96 & 0.07\\
		& GC-RANSAC \cite{barath2018graph} & 67.65 & 2.33 & 6.87 & 48.55 & 69.38 & 56.78 & 92.05 & 2.33 & 7.11 & 64.46 & \textcolor{Red}{93.39} & 75.69 & 0.55\\
		& RANSAC-1M \cite{fischler1981random} & 64.20 & 4.05 & 11.35 & 63.96 & 57.90 & 60.13 & 88.42 & 3.05 & 9.42 & 77.96 & 79.86 & 78.55 & 0.97 \\
		& RANSAC-2M \cite{fischler1981random} & 65.25 & 4.07 & 11.56 & 64.41 & 58.37 & 60.51 & 90.88 & 2.71 & 8.31 & 78.52 & 83.52 & 80.68 & 1.63 \\
		& RANSAC-4M \cite{fischler1981random} & 66.10 & 3.95 & 11.03 & 64.27 & 59.10 & 61.02 & 91.44 & 2.69 & 8.38 & 78.88 & 83.88 & 81.04  & 2.86 \\
		& CG-SAC \cite{quan2020compatibility} & 78.00 & 2.40 & 6.89 & 68.07 & 67.32 & 67.52 & 87.52 & 2.42 & 7.66 & 75.32 & 84.61 & 79.90 & 0.27 \\
		& Ours & \textcolor{Red}{83.98} & 2.18 & \textcolor{Blue}{6.56} & \textcolor{Blue}{72.48} & \textcolor{Red}{78.33} & \textcolor{Red}{75.10} & \textcolor{Red}{93.28} & \textcolor{Red}{2.08} & \textcolor{Blue}{6.55} & 78.94 & \textcolor{Blue}{86.39} & \textcolor{Red}{82.20} & 0.11\\ \hline
		
	\end{tabular}
	\caption{Quantitative results on 3DMatch Dataset. Methods with * are correspondence-free methods. }
	\vspace{-3mm}
	\label{tab:overallPreformance}%
	
\end{table*}

\subsection{Datasets and Experimental Setup}

\textbf{Indoor scenes.} 
We use the 3DMatch benchmark \cite{zeng20173dmatch} for evaluating the performance on indoor scenes. For each pair of point clouds, we first use 5cm voxel grid to down-sample the point cloud. Then we extract the local feature descriptors and match them to form the putative correspondences.  Following \cite{bai2021pointdsc}, we use FPFH \cite{rusu2009fast} (handcrafted descriptor) and FCGF \cite{choy2019fully} (learned descriptor) as feature descriptors respectively. 
The partial overlapping registration benchmark 3DLoMatch \cite{huang2021predator} is also adopted to further verify the performance of the method. Following \cite{lee2021deep,bai2021pointdsc}, we use FCGF \cite{choy2019fully} and Predator \cite{huang2021predator} descriptors to generate putative correspondences.

\textbf{Outdoor scenes.} We use the KITTI dataset \cite{geiger2012we} for testing the effectiveness on outdoor scenes. Following \cite{choy2019fully,choy2020deep}, we choose the 8 to 10 scenes, obtaining 555 pairs of point clouds for testing. Then we construct 30cm voxel grid to down-sampling the point cloud and form the correspondences by FPFH and FCGF descritors respectively.

\textbf{Evaluation Criteria.} Following \cite{bai2021pointdsc}, we first report the registration recall (RR) under an error threshold. For the indoor scenes, the threshold is set to ($15^{\circ}$, 30 cm), while the threshold of outdoor scenes is ($5^{\circ}$, 60 cm). For a pair of point clouds, we calculate the errors of translation and rotation estimation separately. We compute the isotropic rotation error (RE) \cite{ma2012invitation} and L2 translation error (TE) \cite{choy2020deep}. 
Following \cite{bai2021pointdsc}, we also report the outlier removal results using following three evaluation criteria: inlier precision (IP), inlier recall (IP) and F1-measure (F1).

\textbf{Implementation Details.} 
When computing the SC$^2$ matrix, the $d_{thr}$ is set to the twice as the voxel size for down-sampling (10cm for indoor scenes and 60cm for outdoor scenes). The number of seed ($N_{s}$ in Section \ref{step1}) is set to 0.2 * $N$, where $N$ is the number of correspondences. When sampling consensus set, we select 30 nearest neighbors ($K_{1}$ = 30) of seed point at first sampling stage, and remain 20 correspondences ($K_{2}$ = 20) to form the consensus set. All the experiments are conducted on a machine with an INTEL Xeon E5-2620 CPU and a single NVIDIA GTX1080Ti. 

\subsection{Evaluation on Indoor Scenes}
We first report the results on 3DMatch dataset in Tab. \ref{tab:overallPreformance}. We compare our method with 13 baselines: DCP \cite{wang2019deep}, PointNetLK \cite{aoki2019pointnetlk}, 3DRegNet \cite{pais20203dregnet}, DGR \cite{choy2020deep}, DHVR \cite{lee2021deep}, PointDSC \cite{bai2021pointdsc}, SM \cite{leordeanu2005spectral}, ICP \cite{besl1992method}, FGR \cite{zhou2016fast}, TEASER \cite{yang2020teaser}, GC-RANSAC \cite{barath2018graph}, RANSAC \cite{fischler1981random}, CG-SAC \cite{quan2020compatibility}. The first 6 methods are based on deep learning, while the last 7 methods are traditional methods. For the deep learning methods, we use the provided pre-trained model of them for testing. The results of DHVR we tested have some difference with the original results, so we also report the results in their paper (DHVR-Origin in Tab. \ref{tab:overallPreformance}).
DCP, PointNetLK and ICP are correspondence-free methods, so their results are not related with the descriptor.

\begin{table}[t]
	\centering%
	
	\scriptsize
	\renewcommand\tabcolsep{2.8pt}
	\begin{tabular}{c|ccc|ccc|c}
		
		\hline
		& \multicolumn{3}{c|}{FCGF } & \multicolumn{3}{c|}{Predator} \\
		& RR(\%) & RE($\circ$) & TE(cm)  & RR(\%) & RE($\circ$) & TE(cm) & Time(s) \\ \hline
		DHVR \cite{lee2021deep} & 54.41 & 4.14 & 12.56 & 65.41 & 4.97 & 12.33 & 3.55\\
		DGR \cite{choy2020deep} & 43.80 & 4.17 & 10.82 & 59.46 & \textcolor{Red}{3.19} & 10.01 & 1.48\\
		PointDSC \cite{bai2021pointdsc} & \textcolor{Blue}{56.09} & 3.87 & \textcolor{Red}{10.39} & \textcolor{Blue}{68.89} & \textcolor{Blue}{3.43} & \textcolor{Blue}{9.60} & \textcolor{Red}{0.10} \\ \hline
		FGR \cite{zhou2016fast} & 19.99 & 5.28 & 12.98 & 35.99 & 4.77 & 11.64 & 1.32\\
		RANSAC \cite{fischler1981random} & 46.38 & 5.00 & 13.11 & 64.85 & 4.28 & 11.04 & 2.86\\
		CG-SAC \cite{quan2020compatibility} & 52.31 & \textcolor{Blue}{3.84} & 10.55 & 64.01 & 3.86 & 10.94 & 0.25\\
		Ours & \textcolor{Red}{57.83} & \textcolor{Red}{3.77} & \textcolor{Blue}{10.46} & \textcolor{Red}{69.46} & 3.46 & \textcolor{Red}{9.58} & \textcolor{Blue}{0.11}\\
		\hline

	\end{tabular}
	\caption{Quantitative results on 3DLoMatch Dataset. }
	\vspace{-3mm}
	\label{tab:3DLoMatch}%
	
\end{table}

\textbf{Combined with FPFH.} We first use the FPFH descriptor to generate the correspondences, in which the mean inlier rate is 6.84\%. 
As shown in Tab. \ref{tab:overallPreformance}, our method greatly outperforms all the methods. For the registration recall (RR), which is the most important criterion, our method improves it by about 6\% over the closest competitors among the retested results (PointDSC and CG-SAC). Following \cite{choy2020deep,bai2021pointdsc}, since the part of failed registration can generate a large error of translation and rotation, we only compute the mean rotation (RE) and translation error (TE) of successfully registered point cloud pairs of each method to avoid unreliable metrics. This strategy of measurement makes methods with high registration recall more likely to have large mean error, because they include more difficult data when calculating mean error. Nevertheless, our method still achieves competitive results on RE and TE. Our method is slightly worse than PointDSC on TE and RE, and better than other methods. For the outlier rejection results, our method achieves the highest inlier recall (IR) and F1-measure. The F1 of our method outperforms the PointDSC by 5.35\%.

\textbf{Combined with FCGF.} To further verify the performance, we also adopt the recent FCGF descriptor to generate putative correspondences and report the registration results. The mean inlier rate of putative correspondences is 25.61\%. As shown in Tab. \ref{tab:overallPreformance}, since the inlier rate is higher than the correspondences obtained by FPFH descriptor, the performace of all of the feature based methods are boosted.
Our method still achieves the best performance over all the methods, achieving 1.84\% improvement over RANSAC on registration recall. 
Besides, the mean registration time for a pair of point clouds are also reported. Since the proposed method only need to sample a few seed points with their consensus set rather than a large number of samples, it is competitive in terms of time-consuming. As shown in Tab. \ref{tab:overallPreformance}, the mean registration time of our methods is 0.11s, which is over 20$\times$ faster than RANSAC with 4M iterations.

\begin{figure}[t]
	\centering
	\includegraphics[width=1\columnwidth]{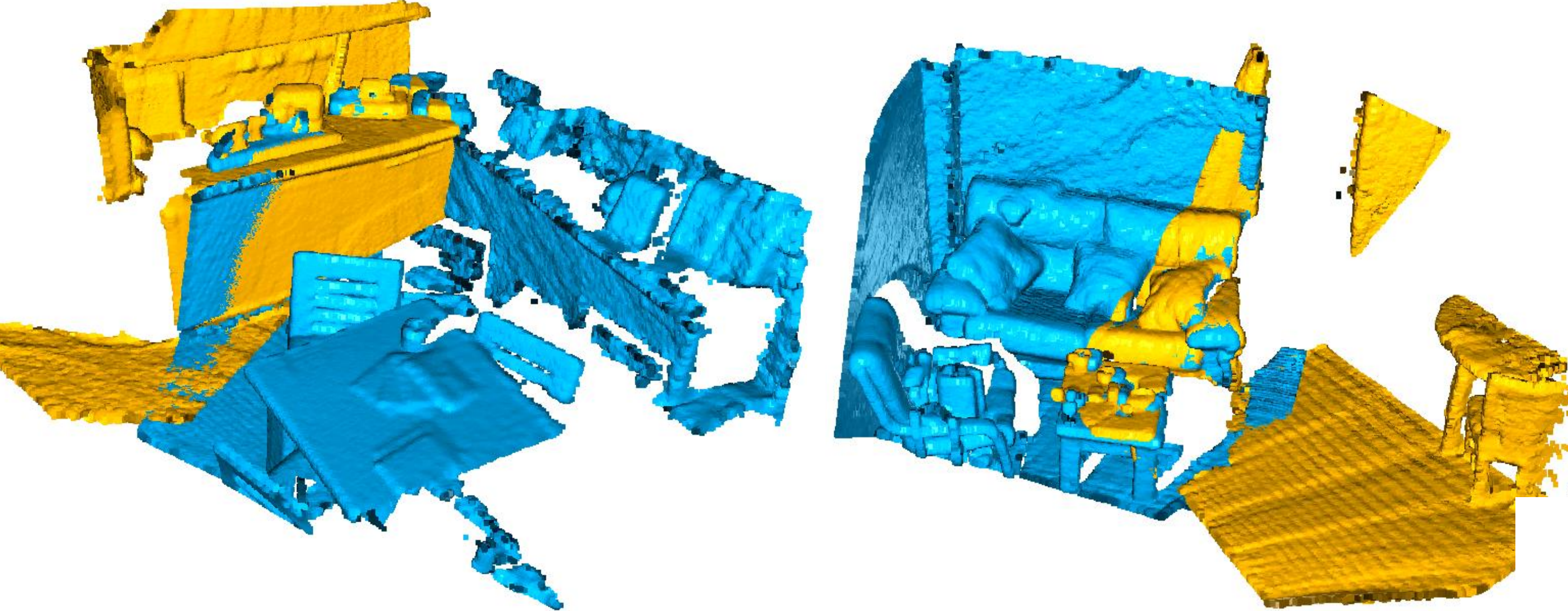}
	\caption{Qualitative registration results on 3DLoMatch dataset.}\label{visual_result}
	\vspace{-3mm}
\end{figure}

\begin{table}[t]
	\centering%
	
	\scriptsize
	\renewcommand\tabcolsep{3.0pt}
	\begin{tabular}{c|ccc|ccc|c}
		
		\hline
		& \multicolumn{3}{c|}{FPFH } & \multicolumn{3}{c|}{FCGF} \\
		& RR(\%) & RE($\circ$) & TE(cm)  & RR(\%) & RE($\circ$) & TE(cm) & Time(s) \\ \hline
		DHVR \cite{lee2021deep}  & - & - & - & \textcolor{Red}{99.10} & \textcolor{Red}{0.29} & \textcolor{Red}{19.80} & 0.83\\
		DGR \cite{choy2020deep} & 77.12 & 1.64 & 33.10 & 96.90 & 0.34 & 21.70 & 2.29\\
		PointDSC \cite{bai2021pointdsc} & \textcolor{Blue}{98.20} & \textcolor{Blue}{0.35} & \textcolor{Blue}{8.13} & 98.02  & \textcolor{Blue}{0.33} & 21.03 & \textcolor{Blue}{0.45}\\
		 \hline
		FGR \cite{zhou2016fast} & 5.23 & 0.86 & 43.84 & 89.54 & 0.46 & 25.72 & 3.88\\
		RANSAC \cite{fischler1981random} & 74.41 & 1.55 & 30.20 & 80.36 & 0.73 & 26.79 & 5.43\\ 
		CG-SAC \cite{quan2020compatibility} & 74.23 & 0.73 & 14.02 & 83.24 & 0.56 & 22.96 & 0.73\\
		Ours & \textcolor{Red}{99.64} & \textcolor{Red}{0.32} & \textcolor{Red}{7.23} & \textcolor{Blue}{98.20} & \textcolor{Blue}{0.33} & \textcolor{Blue}{20.95} & \textcolor{Red}{0.31}\\
		\hline
	\end{tabular}
	\caption{Quantitative results on KITTI Dataset. (For DHVR, the authors have neither released training code nor pretraining model on KITTI dataset, so we report the results in their paper.)} \label{tab:KITTI}%
	\vspace{-4mm}
\end{table}

\textbf{Robustness to lower overlap.} Furthermore, we report the results on the low-overlap scenarios: 3DLoMatch \cite{huang2021predator}. Following PointDSC \cite{bai2021pointdsc} and DHVR \cite{lee2021deep}, we adopt the FCGF \cite{choy2019fully} and Predator (There are two versions of Predator, and we use the updated one) \cite{huang2021predator} descriptors to generate correspondences. The registration recall (RR), rotation error (RE) and translation error (TE) are reported in Tab. \ref{tab:3DLoMatch}. As shown by the data, 
whether combined with FCGF or Predator descriptor, our method achieves the highest registration recall. Meanwhile, we also present some qualitative results on 3DLoMatch dataset. As shown in Fig. \ref{visual_result}, our method can successfully align two point clouds where the low overlap ratio is clearly visible.

\subsection{Evaluation on Outdoor Scenes}
In this experiments, we test on the outdoor KITTI \cite{geiger2012we} dataset. The results of DHVR \cite{lee2021deep}, DGR \cite{choy2020deep}, PointDSC \cite{bai2021pointdsc}, RANSAC \cite{fischler1981random}, FGR \cite{zhou2016fast}, CG-SAC \cite{quan2020compatibility} are reported as comparison. DHVR, DGR and PointDSC are deep learning based methods, while the remaining methods are non-learning. As shown in Tab. \ref{tab:KITTI}, our method remarkably surpasses the non-learning methods. The registration recall (RR) of our method is 25.23\% higher than that of RANSAC when combined with FPFH descriptor, and 17.84\% higher when combined with FCGF descriptor. The errors of translation and rotation are also much lower than RANSAC. Our method with FPFH descriptor obtains the results of highest registration recall. For the learning networks, our method can achieve close performance with them.

\subsection{Generalization and Robustness}
\textbf{Generalization experiments.} As reported above, deep learning based methods also achieve competitive performance on the 3DMatch, 3DLoMatch and KITTI datasets. Compared with these methods based on deep learning, the other advantage of our method is that it has no bias cross different datasets, while deep learning based methods have performance degradation when generalized between different datasets. To demonstrate this, we perform the generalization experiments on both 3DMatch, 3DLoMatch and KITTI dataset. For the recent learning based methods, including DGR and PointDSC, we report the cross-dataset results. Specifically, we adopt their pre-trained model by KITTI to test on 3DMatch and 3DLoMatch and use 3DMatch's model to test on KITTI. As shown in Tab. \ref{Generalization}, our method shows a significant improvement on registration recall without generalization problem. This further demonstrates the effectiveness of our method.

\begin{table}[t]
	\centering%
	
	\scriptsize
	\renewcommand\tabcolsep{3.0pt}
	\begin{tabular}{c|cc|cc|cc}
		
		\hline
		& \multicolumn{2}{c|}{3DMatch } & \multicolumn{2}{c|}{3DLoMatch} & \multicolumn{2}{c}{KITTI} \\
		& FPFH & FCGF & FCGF & Predator & FPFH  & FCGF\\ \hline
		DGR-gen & 49.48 & 81.89 & 23.75 & 45.03 & 73.69 & 86.12 \\
		PointDSC-gen & 68.12 & 87.74 & 40.65 & 53.79 & 90.27 & 92.97  \\
		Ours & 83.98 & 93.28 & 57.83 & 69.46 & 99.64 & 98.20 \\
		\hline
	\end{tabular}
	\caption{Generalization results. The registration recall (\%) on 3DMatch, 3DLoMatch and KITTI datasets are reported. } \label{Generalization}%
	\vspace{-2mm}
\end{table}

\begin{figure}[t]
	\centering
	\includegraphics[width=1\columnwidth]{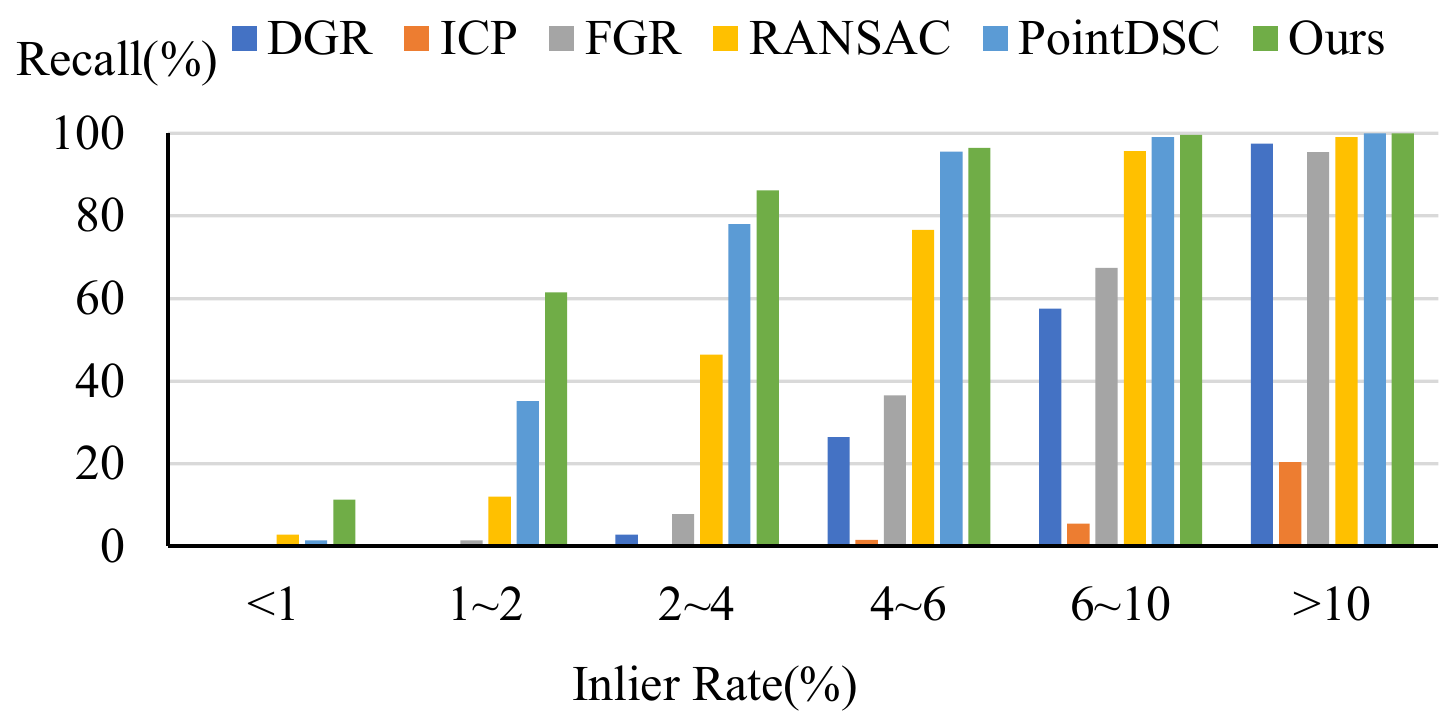}
	\caption{The registration recall under the different inlier ratio of the putative correspondences.}\label{inlier_rate}
	
	\vspace{-5mm}
\end{figure}

\textbf{Robustness Anti Noise.}
An important factor to measure the performance of the model fitting method is the stability under low inlier rate. In order to further verify the performance of our method, we report the results under different inlier ratio in Fig. \ref{inlier_rate}. Specifically, we first use FPFH to generate initial match pairs for the 3DMatch dataset. Then, according to the inlier ratio, all the point cloud pairs are divided into 6 groups: $<$ 1\%, 1\% - 2\%, 2\% - 4\%, 4\% - 6\%, 6\% - 10\% and $>$ 10\%. The number of point cloud pairs in each group is: 141, 208, 346, 252, 323 and 353. As shown in Fig. \ref{inlier_rate}, when the inlier rate is less than 2\%, our method is significantly better than other methods. It demonstrates the robustness anti noise of our method.

\subsection{Combined with learning network}
To verify the flexibility of our proposed approach, we combine our approach with a recent deep learning approach PointDSC \cite{bai2021pointdsc}. It adopts the spatial consistency matrix to guide the non-local module. Since the proposed SC$^2$ measure is more robust to the ambiguity, we replace the spatial consistency matrix in PointDSC with SC$^2$. Instead of retraining the network, we directly plugged our metrics into it. The registration recall of their vanilla version and combined version are shown in Tab. \ref{tab:combine}. It can be seen that adding our metrics can significantly boost the performance of network, especially for the generalization performance of network. It demonstrates that the proposed measure is flexible to combine with other methods.

\subsection{Ablation Study}
In this section, we perform ablation study on 3DMatch dataset. We use the FPFH and FCGF descriptors to form correspondences respectively. The classic RANSAC is adopted as our baseline, as shown in the Row 1 and Row 7 of Tab. \ref{tab:Ablation}. We progressively add the proposed modules to the baseline and report the results.  

\textbf{Second Order Spatial Compatibility.} We first add the Second Order Spatial Compatibility (SC$^2$) measure as the guidance for the sampling of RANSAC. Each correspondence is extended into a consensus set by seaching the $k$-nearest neighbors in metric space. The Spatial Compatibility (SC) adopted by previous works \cite{bai2021pointdsc,quan2020compatibility,yang2021sac} is also utilized as the sampling guidance, and the results are reported as comparison. As shown in Row 1, 3 and 7, 9 of Tab. \ref{tab:Ablation}, the registration recall obtained by using SC$^2$ measure as guidance is 14.79\% higher than RANSAC when combined with FPFH, and 1.66\% higher when combined with FCGF. Meanwhile, since SC$^2$ measure can narrow the sampling space, the mean registration time of SC$^2$ measure is much smaller than RANSAC. Besides, using SC$^2$ measure as guidance can achieve better performance than using SC measure by comparing Row 2, 3 and 8, 9. This is because SC is disturbed by the ambiguity problem, while SC$^2$ measure can eliminate the ambiguity.

\textbf{Two-stage Selection.} We further adopt a two-stage selection strategy for generating consensus set for each seed. When one seed is an inlier correspondence, it has almost removed most of the outliers in the consensus set formed in the first stage. Since SC$^2$ becomes more stable when the inlier rate increases, we construct a local SC$^2$ matrix to remove potential outliers. Comparing Row 3, 4 and 9, 10 in Tab. \ref{tab:Ablation}, using two-stage selection achieves a recall improvement of 1.96\% when combined with FPFH, and 0.12\% improvement when combined with FCGF.

\begin{table}[t]
	\centering%
	
	\scriptsize
	\renewcommand\tabcolsep{3.0pt}
	\begin{tabular}{c|cc|cc|cc}
		
		\hline
		& \multicolumn{2}{c|}{3DMatch } & \multicolumn{2}{c|}{3DLoMatch} & \multicolumn{2}{c}{KITTI} \\
		& FPFH & FCGF & FCGF & Predator & FPFH  & FCGF\\ \hline
		PointDSC & 77.57 & 92.85 & 56.09 & 68.89  & 98.20 & 98.02 \\
		PointDSC+SC$^2$ & 83.24 & 93.10 & 57.05 & 69.04 & 99.10 & 98.02 \\
		PointDSC-gen & 68.12 & 87.74 & 40.65 & 53.79 & 90.27 & 92.97 \\
		PointDSC-gen+SC$^2$ & 74.12 & 89.59 & 44.81 & 56.71 & 97.48 & 98.02 \\
		\hline
	\end{tabular}
	\caption{The registration recall of combining SC$^2$ measure with learning based network. } \label{tab:combine}%
	\vspace{-3mm}
\end{table}

\textbf{Local Spectral Matching.} When a minimum set is sampled, RANSAC adopt the instance-equal SVD to generate an estimation of translation and rotation, which is sensitive to errors. We replace the instance-equal SVD \cite{arun1987least} with the weighted SVD \cite{pais20203dregnet,choy2020deep}, so that less reliable correspondences are assigned lower weights for robust registration. We construct a soft SC$^2$ matrix in each consensus set, and then use local spectral matching to compute the association between each correspondence with the main cluster. The association value is utilized as the weight for weighted SVD. Comparing Row 4, 5 and 10, 11 in Tab. \ref{tab:Ablation}, using local spectral matching can boost the performance, especially for the mean rotation and translation error.

\textbf{Seed Selection.} So far, each correspondence is treated as a seed. However, it do not need to generate a consensus set for all correspondences and estimate a rigid transformation. We only need to select a few reliable points, and use the aggregation among the inliers to collect the set without outliers, so as to further improve the efficiency of registration. We use the global spectral matching combined with Non-Maximum Suppression to find several correspondences as seeds instead of all set. Row 5, 6 and 11, 12 of Tab. \ref{tab:Ablation} shows that Seed Selection can reduce registration time by more than half without much performance degradation.

\begin{table}[t]
	\centering%
	\scriptsize
	\renewcommand\tabcolsep{3.0pt}

	\begin{tabular}{cc|ccccc|cccc}
		\hline 
		& & SC & SC$^2$ & TS & LSM & Seed & RR(\%) & RE($\circ$) & TE(cm) & Time(s)  \\ \hline 
		\multirow{6}{0.2cm}{\rotatebox{90}{FPFH}} & 1) & & & & & & 66.10 & 3.95 & 11.03 & 2.86\\ 
		& 2) & \checkmark &  & & & & 71.56 & 2.07 & 6.48 & 0.27\\ 
		
		& 3) & & \checkmark & & & & 80.89 & 2.34 & 6.92 & 0.31\\ 
		& 4) & & \checkmark & \checkmark & & &82.85 & 2.32 & 6.69 & 0.33 \\
		& 5) & & \checkmark & \checkmark & \checkmark &  & 84.10 & 2.13 & 6.56 & 0.37 \\ 
		& 6) & & \checkmark & \checkmark & \checkmark & \checkmark & 83.98 & 2.18 & 6.56 & 0.11 \\  \hline

		\multirow{6}{0.2cm}{\rotatebox{90}{FCGF}} & 7) & & & & & & 91.44 & 2.69 & 8.38  & 2.86\\
		& 8) & \checkmark & & & & & 87.52 & 2.42 & 7.66 & 0.27\\ 
		& 9) & & \checkmark & & &  & 93.10 & 2.16 & 6.76 & 0.31 \\ 
		& 10) & & \checkmark & \checkmark & & & 93.22 & 2.10 & 6.88 & 0.33 \\  
		& 11) & & \checkmark & \checkmark & \checkmark & & 93.28 & 2.08 & 6.56 & 0.37  \\ 
		& 12) & & \checkmark & \checkmark & \checkmark & \checkmark & 93.28 & 2.08 & 6.55 & 0.11 \\  \hline

		\hline 
	\end{tabular}
	
	\caption{Ablation study on 3DMatch Dataset. \textbf{SC}: Spatial compatibility measure. \textbf{SC$^2$}: Second Order Spatial compatibility measure. \textbf{TS}: Two-stage Selection for consensus set sampling. \textbf{LSM} Local Spectral Matching. \textbf{Seed}: Using seed points to reduce the number of sampling.
	} %
	\label{tab:Ablation}
	\vspace{-5mm}
\end{table}

\section{Conclusion}
In this paper, we present a second order spatial compatibility (SC$^2$) measure based point cloud registration method, called SC$^2$-PCR. The core component of our method is to cluster inliers by the proposed SC$^2$ measure at early stage while eliminating ambiguity. Specifically, some reliable correspondences are selected by a global spectral decomposition with Non-Maximum Suppression firstly, called seed points. Then a two-stage sampling strategy is adopted to extend the seed points into some consensus sets. After that, each consensus set produces a rigid transformation by local spectral matching. Finally, the best estimation is selected as the final result. Extensive experiments demonstrate that our method achieves the state-of-the-art performance and high efficiency. Meanwhile, we also demonstrate the proposed SC$^2$ is a flexible measure, which can be combined with learning networks to further boost their performance.

\section{Acknowledgements}
This work was supported by the National Natural Science
Foundation of China under Grants 62176096, 62176242 and 61991412.

{\small
\bibliographystyle{ieee_fullname}
\bibliography{egbib}
}

\clearpage
{ 
\setcounter{page}{1}
\section{Supplementary Material}
\subsection{Derivations}
\textbf{Derivation of Eq. \ref{Ambiguity_SC2}.} We first reformulate Eq. (\ref{global_compatibility}) as follows:
\begin{equation}
	\begin{aligned}
		SC^{2}_{ij} = C_{ij} \cdot M_{ij},
	\end{aligned}\label{rewritten_SC2}
\end{equation}
where $M_{ij}$ is computed as follows:
\begin{equation}
	\begin{aligned}
		M_{ij} = \sum_{k=1}^{N}C_{ik} \cdot C_{kj}.
	\end{aligned}\label{rewritten_ambiguity_tmp}
\end{equation}
$M_{ij}$ counts the quantity of the commonly compatible correspondences of $i$ and $j$ in the global set.
According to Eq. (\ref{compatibility_matrix}) and (\ref{PDF_in_in}), we can obtain that:
\begin{equation}
	\begin{aligned}
		{\rm P} (C_{in,in} = 1) = 1.
	\end{aligned}\label{P_C_inin}
\end{equation} 
According to Eq. (\ref{compatibility_matrix}), (\ref{PDF_in_out}) and (\ref{F0}), we can get that
\begin{equation}
	\begin{aligned}
		{\rm P} (C_{in,out} = 1) = \int_{0}^{d_{thr}} F(l)dl = d_{thr} \cdot f_0 = p
	\end{aligned}\label{P_C_inout} 
\end{equation} 
\begin{equation}
	\begin{aligned}
		{\rm P} (C_{out,out} = 1) = \int_{0}^{d_{thr}} F(l)dl = d_{thr} \cdot f_0 = p.
	\end{aligned}\label{P_C_outout}
\end{equation} 
According to Eq. (\ref{rewritten_SC2}), to make $SC^{2}_{in,out} > SC^{2}_{in,in}$ hold, two conditions need to be met: $C_{in,out}=1$ and $M_{in,out} > M_{in,in}$. According to Eq. (\ref{P_C_inout}), we can obtain the following equation:
\begin{equation}
	\begin{aligned}
		&{\rm P}(SC^{2}_{in,out} > SC^{2}_{in,in}) 
		\\ =&{\rm P}(C_{in,out} = 1) \cdot {\rm P}  (M_{in,out} > M_{in,in}) \\
		=& p \cdot {\rm P}  (M_{in,out} > M_{in,in}).
	\end{aligned}\label{P_ambiguity2}
\end{equation} 

Next, we compute the distribution of $M_{in,out}$ and $M_{in,in}$. Since inliers have different distribution with outliers, we consider them separately and reformulate Eq. (\ref{rewritten_ambiguity_tmp}) as follows:
\begin{equation}
	\begin{aligned}
		M_{ij} = \sum\limits_{m \in \mathcal{I}}C_{im} \cdot C_{mj} + \sum\limits_{n \in \mathcal{O}}C_{in} \cdot C_{nj}.
	\end{aligned}\label{rewritten_ambiguity}
\end{equation}
where $\mathcal{I}$ is the inlier set while $\mathcal{O}$ is the outlier set. (For conveniecne we use this notation in the following part).

We first discuss the value in $M$ matrix between two inliers, i.e. $M_{in,in}$.
According to Eq. (\ref{P_C_inin}), we can find that any two inliers are compatible. Thus, when correspondence $i$ and $j$ are inliers, the number of correspondences compatible with both of them in the inlier set is the number of inliers excluding themselves ($C_{ii}=0, C_{jj}=0$), i.e.:
\begin{equation}
	\begin{aligned}
		\sum\limits_{m \in \mathcal{I}}C_{im} \cdot C_{mj} = N \cdot \alpha - 2; i \in \mathcal{I}, j \in \mathcal{I},
	\end{aligned}
\end{equation}
where $\alpha$ is the inlier rate.
For outliers, according to Eq. (\ref{P_C_inout}), the probability that an outlier is compatible with an inlier is $p$. Then the probability that an outlier is compatible with both $i$ and $j$ is $p^{2}$. The number of outliers in the whole correspondence set is $N(1 - \alpha)$. So the number of correspondences compatible with both of them in the outlier set is in a Bernoulli distribution \cite{box1978statistics} as follows:
\begin{equation}
	\begin{aligned}
		\sum\limits_{n \in \mathcal{O}}C_{in} \cdot C_{nj} \sim B(N(1 - \alpha), p^{2}); i \in \mathcal{I}, j \in \mathcal{I},
	\end{aligned}
\end{equation}
where $B(\cdot, \cdot)$ is the Bernoulli distribution.
Thus, ${M}_{in,in}$ is in the following distribution:
\begin{equation}
	\begin{aligned}
		M_{in,in} \sim N \cdot \alpha - 2 + B(N(1 - \alpha), p^{2}).
	\end{aligned}\label{M_inin}
\end{equation}

After that, we discuss the distribution of the value in $M$ matrix between an inlier and an outlier, i.e. $M_{in,out}$. For convenience, we assume correspondence $i$ is inlier while $j$ is outlier. For the inlier set except correspondence $i$ ($C_{ii}=0$), any of them is compatible with $i$ (Eq. \ref{P_C_inin}), and the probability that one of them is compatible with $j$ is $p$ (Eq. \ref{P_C_inout}). So the number of correspondences compatible with both of correspondence $i$ and $j$ in the inlier set is in following distribution:
\begin{equation}
	\begin{aligned}
		\sum\limits_{m \in \mathcal{I}}C_{im} \cdot C_{mj} \sim  B(N \alpha - 1, p); i \in \mathcal{I}, j \in \mathcal{O}.
	\end{aligned}
\end{equation}
Meanwhile, for each outlier except correspondence $j$ ($C_{jj} = 0$), the probabilities that it is compatible with $i$ or $j$ are both $p$ (Eq. \ref{P_C_inout} and \ref{P_C_outout}). So the probability that an outlier is both compatible with $i$ and $j$ is $p^{2}$. Thus, we can get the following distribution:
\begin{equation}
	\begin{aligned}
		\sum\limits_{n \in \mathcal{O}}C_{in} \cdot C_{nj} \sim B(N(1 - \alpha) - 1, p^{2}); i \in \mathcal{I}, j \in \mathcal{O}.
	\end{aligned}
\end{equation}
So the distribution of $M_{in,out}$ is as follows:
\begin{equation}
	\begin{aligned}
		M_{in,out} \sim B(N \alpha - 1, p) + B(N(1 - \alpha) - 1, p^{2}); 
	\end{aligned}\label{M_inout}
\end{equation}
Since $p$ is a small value, the Binomial distribution in Eq. (\ref{M_inin}) and (\ref{M_inout}) can be approximately equivalent to the Poisson distribution \cite{box1978statistics}, i.e.:
\begin{equation}
	\begin{aligned}
		M_{in,in} &\sim N \cdot \alpha - 2 + \pi(N(1 - \alpha) p^{2}), \\
		M_{in,out} &\sim \pi ((N \alpha - 1)p) + \pi((N(1 - \alpha) - 1) p^{2}),
	\end{aligned}\label{M}
\end{equation}
where $\pi(\cdot)$ is the Poisson distribution.
Furthermore, for two Poisson distribution: $X_1 \sim \pi(\lambda_{1})$ and $X_2 \sim \pi(\lambda_{2})$, their sum is also in the Poisson distribution \cite{box1978statistics} as follows:
\begin{equation}
	\begin{aligned}
		X1 + X_2 \sim \pi(\lambda_{1} + \lambda_{2}).
	\end{aligned}
\end{equation}
So we can convert $M_{in,out}$ in Eq. \ref{M} into following form:
\begin{equation}
	\begin{aligned}
		M_{in,out} \sim \pi ((N \alpha - 1)p + (N(1 - \alpha) - 1) p^{2}).
	\end{aligned}\label{M_inout2}
\end{equation}
Meanwhile, we can convert ${\rm P} (M_{in,out} > M_{in,in})$ into following form: 
\begin{equation}
	\begin{aligned}
		&{\rm P} (M_{in,out} > M_{in,in}) \\
		=&{\rm P} (M_{in,out} - M_{in,in} > 0) \\
		=&{\rm P} (X > N \cdot \alpha - 2),
	\end{aligned}\label{P_M_inout}
\end{equation}
where $X$ is in following distribution:
\begin{equation}
	\begin{aligned}
		X \sim \pi ((N \alpha - 1)p + (N(1 - \alpha) - 1) p^{2})
		- \pi(N(1 - \alpha)p^{2})
	\end{aligned}\label{X_distribution}
\end{equation}
For two Poisson distribution: $X_1 \sim \pi(\lambda_{1})$ and $X_2 \sim \pi(\lambda_{2})$, their difference is in the Skellam distribution \cite{irwin1937frequency,karlis2003analysis,karlis2006bayesian}, i.e.:
\begin{equation}
	\begin{aligned}
		X1 - X_2 \sim S(\lambda_{1}, \lambda_{2}).
	\end{aligned}
\end{equation}
So the distribution of $X$ in Eq. (\ref{X_distribution}) can be converted as follows:
\begin{equation}
	\begin{aligned}
		S((N \alpha - 1) p + (N (1 &- \alpha) - 1) p^{2}, N (1 - \alpha) p^{2}).
	\end{aligned}\label{distribution_X_final}
\end{equation}
Combining Eq. (\ref{P_ambiguity2}), (\ref{P_M_inout}) and (\ref{distribution_X_final}), we compute the value of ${\rm P}(SC^{2}_{in,out} > SC^{2}_{in,in})$ as Eq. (\ref{Ambiguity_SC2}).

\subsection{Additional Experiments}
\textbf{Parameter $K_1$ and $K_2$.}
In the proposed method, when some seeds are selected, we use a two-stage selection strategy to extend each seed into a consensus set. In the first stage, $K_1$ correspondences are selected by finding top-$K_{1}$ neighbors of seed. In the second stage, a local SC$^2$ matrix is rebuilt to further filter potential outliers and reserve $K_2$ correspondences. In order to show the effect of these two parameters, we report the registration results with respect to these parameters in Tab. \ref{Parameter_k}. 

\begin{table}[!htbp]
	\centering%
	
	\footnotesize
	\renewcommand\tabcolsep{2.8pt}
	\begin{tabular}{cc|ccc|ccc|c}
		
		\hline
		\multicolumn{2}{c|}{Param} & \multicolumn{3}{c|}{FPFH } & \multicolumn{3}{c|}{FCGF} \\
		$K_1$ & $K_2$ & RR(\%) & RE($\circ$) & TE(cm)  & RR(\%) & RE($\circ$) & TE(cm) & Time(s)\\ \hline
		10 & 3   & 82.13 & 2.15 & 6.67 & 92.91 & 2.05 & 6.52 & 0.11 \\
		10 & 5   & 82.69 & 2.10 & 6.59 & 93.04 & 2.05 & 6.52 & 0.11 \\
		20 & 10 &  83.79 & 2.13 & 6.56 & 93.10 & 2.06 & 6.54 & 0.11 \\
		30 & 20   & 83.98 & 2.18 & 6.56 & 93.28 & 2.08 & 6.55 & 0.11\\
		40 & 30  & 83.67 & 2.15 & 6.71 & 93.22 & 2.05 & 6.52 & 0.12\\
		50 & 40   & 83.67 & 2.16 & 6.78 & 93.16 & 2.06 & 6.54 & 0.12\\
		60 & 50   & 83.92 & 2.18 & 6.68 & 93.10 & 2.05 & 6.49& 0.12 \\
		70 & 60  & 83.55 & 2.16 & 6.69  & 92.98 & 2.04 & 6.49 & 0.13\\
		\hline
	\end{tabular}
	\caption{The registration results with varying parameters on 3DMatch dataset.}
	\vspace{-3mm}
	\label{Parameter_k}%
	
\end{table}

As shown in Tab. \ref{Parameter_k}, our method is parameter insensitive. The proposed method with different parameters can all lead to acceptable results combined with both FPFH and FCGF descriptors. In the final version, we choose $K_{1}$=30 and $K_{2}$=20 for its best registration recall.

\textbf{Heat Map.} Compared with the spatial compatibility (SC) measure \cite{quan2020compatibility,bai2021pointdsc}, the second order spatial compatibility (SC$^2$) measure can reduce the probability of ambiguity event. In order to show the difference between these two measures, we report the heat maps of them on a real data in Fig. \ref{heapmap}. Specifically, we sort all the correspondences, placing inliers first and outliers behind to build the SC and SC$^2$ matrices respectively. Then we normalize 
these two matrices to [0,1], and plot the corresponding heat maps. As shown in Fig. \ref{heapmap}, there are high values on the right and bottom in the heat map of SC measure, showing that many outliers are
compatible with inliers. By contrast, the right and bottom sides are clean in the heat map of SC$^2$ measure, showing that outliers have low compatibility with inliers.
\begin{figure}[t]
	\centering
	\includegraphics[width=1\columnwidth]{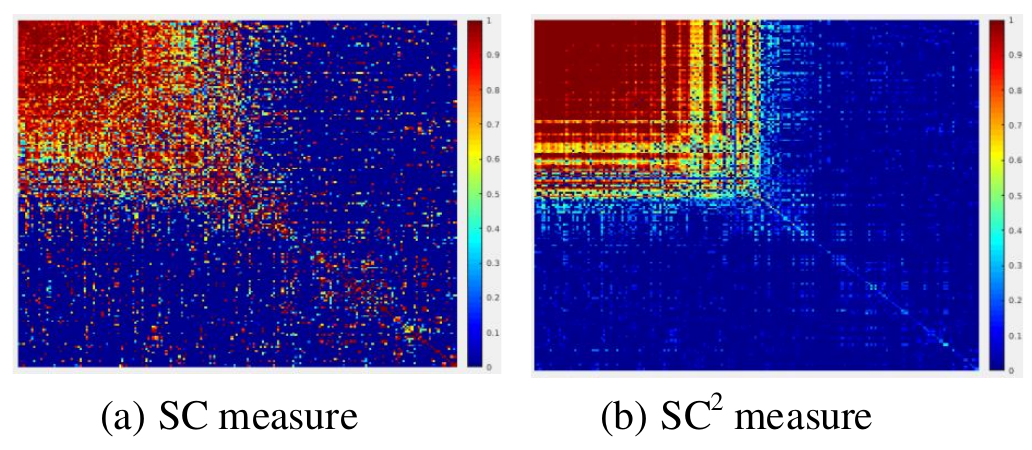}
	\caption{The heat maps of spatial compatibility (SC) and second order spatial compatibility (SC$^2$) measures.}\label{heapmap}
	\vspace{-5mm}
\end{figure}

\textbf{Scene-wise Results.} Following \cite{bai2021pointdsc,choy2020deep}, we also report the scene-wise registration results of the proposed method on 3DMatch dataset, combing both FPFH and FCGF descriptors as shown in Tab. \ref{Scene_wise}.

\begin{table}[!htbp]
	\centering%
	
	\small
	\renewcommand\tabcolsep{2.8pt}
	\begin{tabular}{c|ccc|ccc}
		
		\hline
		& \multicolumn{3}{c|}{FPFH } & \multicolumn{3}{c}{FCGF} \\
		& RR(\%) & RE($\circ$) & TE(cm)  & RR(\%) & RE($\circ$) & TE(cm) \\ \hline
		Kitchen  & 88.34 & 1.95 & 5.41 & 99.21 & 1.69 & 5.15\\
		Home1  & 89.74 & 1.82 & 6.23 & 96.79 & 1.79 & 6.37\\
		Home2  & 73.56 & 2.80 & 7.46 & 83.17 & 3.48 & 7.50 \\
		Hotel1  & 92.04 & 2.20 & 7.01 & 98.67 & 1.89 & 6.08\\
		Hotel2  & 81.73 & 2.08 & 6.49 & 91.35 & 1.94 & 5.61\\
		Hotel3  & 90.74 & 2.01 & 5.79 & 92.59 & 2.08 & 5.80\\
		Study  & 76.03 & 2.31 & 8.94 & 88.36 & 2.31 & 9.21\\
		Lab &  76.62 & 1.67 & 5.99 & 80.52 & 1.91 & 8.44 \\
		\hline

	\end{tabular}
	\caption{Scene-wise registration results on 3DMatch Dataset. }
	\vspace{-3mm}
	\label{Scene_wise}%
	
\end{table}

\textbf{More Qualitative Results.} We show the more registration results in Fig. \ref{3DMatch_result} and \ref{KITTI_result}.

\begin{figure*}[t]
	\centering
	\includegraphics[width=2\columnwidth]{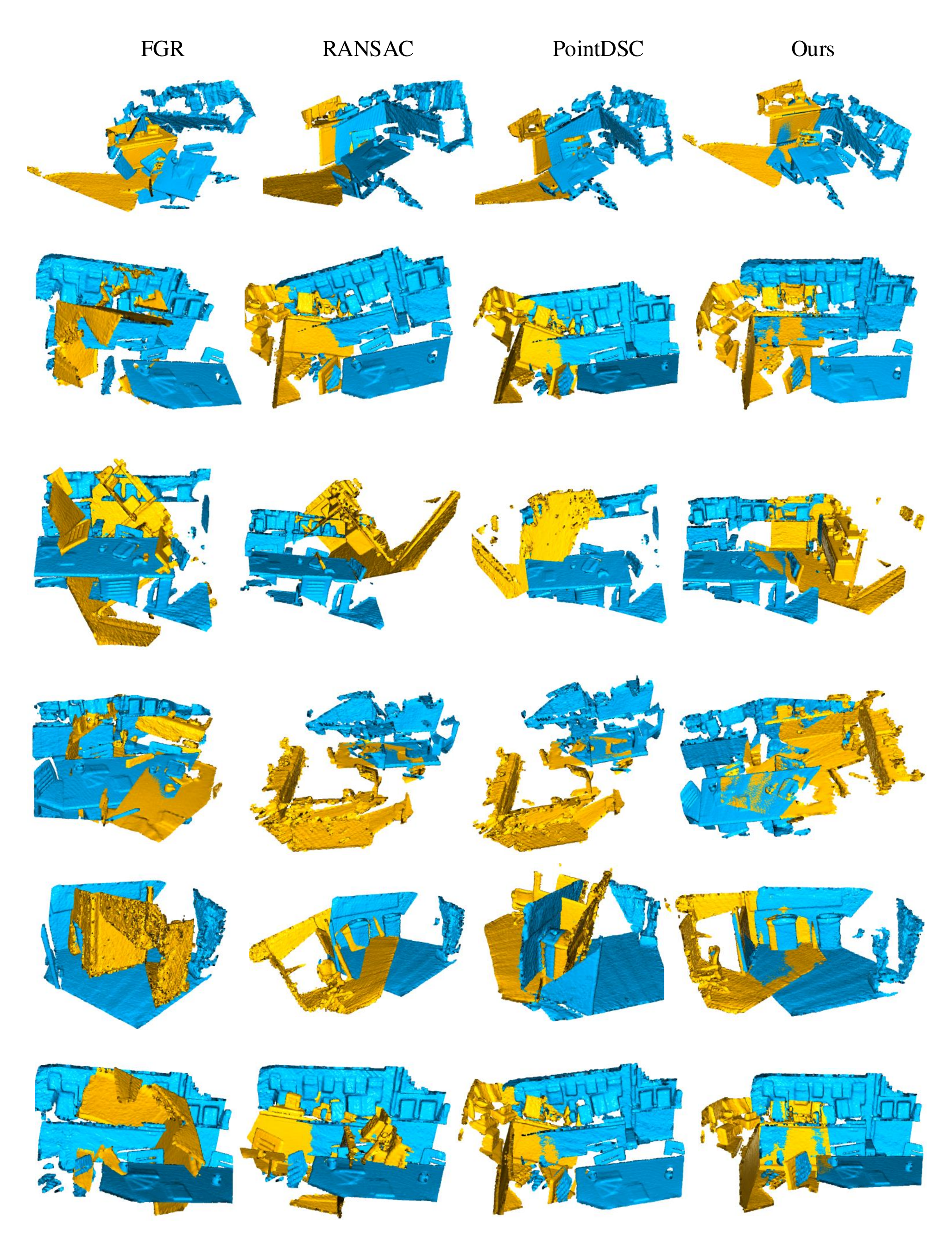}
	\caption{Qualitative comparison on 3DMatch and 3DLoMatch dataset. From left to right are: FGR \cite{zhou2016fast}, RANSAC \cite{fischler1981random}, PointDSC \cite{bai2021pointdsc} and Ours}\label{3DMatch_result}
\end{figure*}

\begin{figure*}[t]
	\centering
	\includegraphics[width=2\columnwidth]{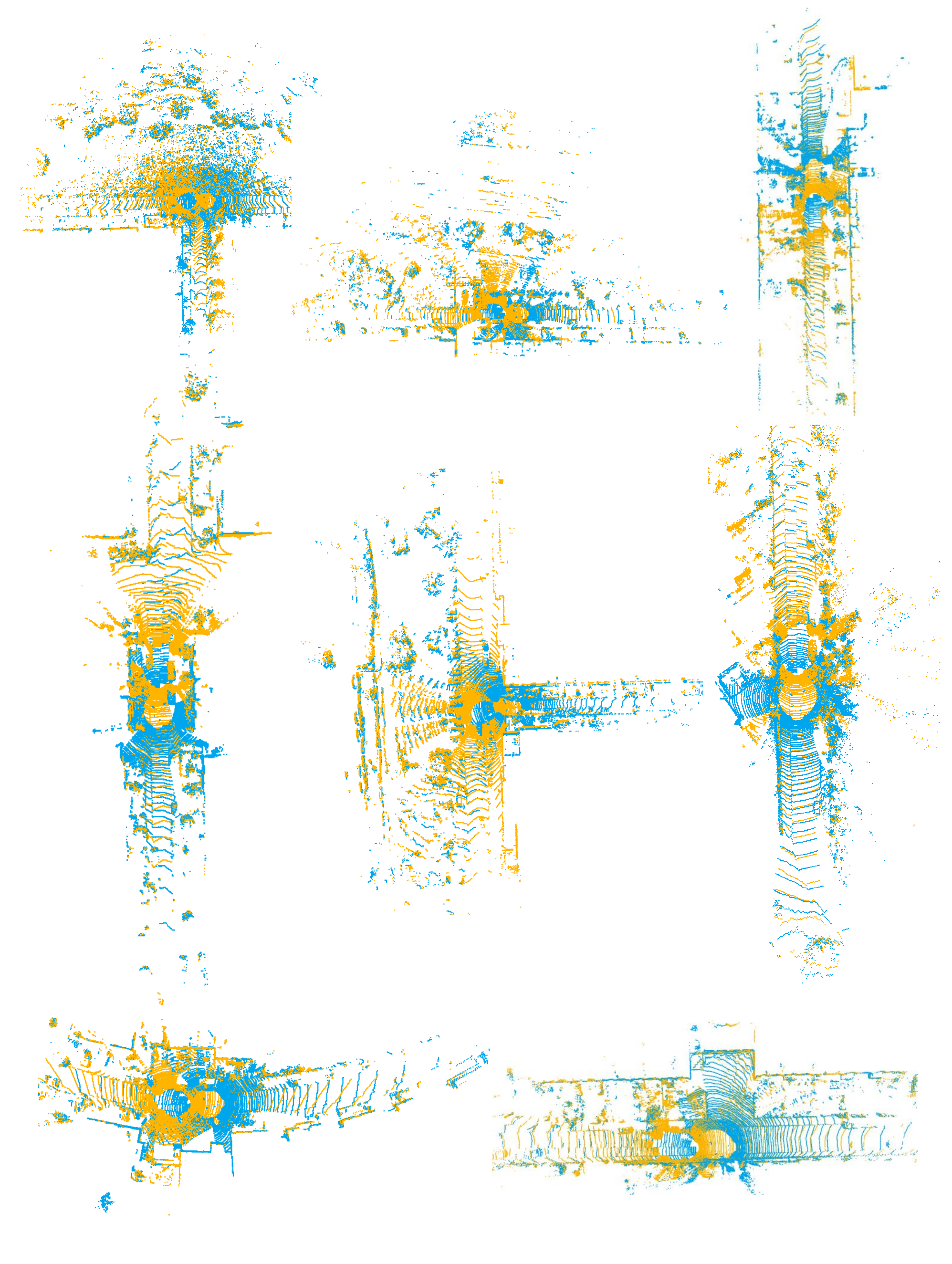}
	\caption{Qualitative results of our method on KITTI dataset. }\label{KITTI_result}
\end{figure*}
	
}

\end{document}